\documentclass[runningheads]{llncs}

\usepackage[final,ID=1128,year=2026]{eccv}

\usepackage{eccvabbrv}

\usepackage{graphicx}
\usepackage{booktabs}
\usepackage{multirow}
\usepackage{multicol}
\usepackage{bm}

\usepackage[accsupp]{axessibility}  
\usepackage{orcidlink}
\usepackage{colortbl}
\usepackage{ulem}
\usepackage{tikz}
\usepackage{pgfplots}
\usepgfplotslibrary{groupplots}
\pgfplotsset{compat=1.18}

\usepackage{hyperref}

\usepackage{orcidlink}

\begin{document}

\title{Dual-Anchoring: Addressing State Drift in Vision-Language Navigation}



\makeatletter
\def\@fnsymbol#1{%
  \ensuremath{%
  \ifcase#1\or \dagger\or \ddagger\or \mathsection\or *\else \@ctrerr\fi}}
\makeatother

\author{Kangyi Wu\inst{1}\thanks{Equal contribution.} \and
Pengna Li\inst{1}\textsuperscript{\ensuremath{\dagger}} \and
Kailin Lyu\inst{2}\thanks{These authors contributed equally.} \and
Xi Lin\inst{3}\textsuperscript{\ensuremath{\ddagger}} \and
Lin Zhao\inst{4}\thanks{Project Leader.} \and
Qingrong He\inst{4} \and
Jinjun Wang\inst{1}\inst{5} \and
Jianyi Liu\inst{1}\thanks{Corresponding author.}}

\authorrunning{K.~Wu et al.}

\institute{National Key Laboratory of Human-Machine Hybrid Augmented Intelligence,\\
National Engineering Research Center for Visual Information and Applications,\\
Institute of Artificial Intelligence and Robotics, Xi'an Jiaotong University \and
Institute of Automation, Chinese Academy of Sciences,\\
School of Artificial Intelligence, University of Chinese Academy of Sciences \and
Johns Hopkins University \and
Joy Future Academy, JD \and
Xpeng Inc.}

\maketitle

\vspace{-0.5cm}
\begin{abstract}
Vision-Language Navigation~(VLN) requires an agent to navigate through 3D environments by following natural language instructions. 
While recent Video Large Language Models~(Video-LLMs) have largely advanced VLN, they remain highly susceptible to \textit{State Drift} in long scenarios. 
In these cases, the agent's internal state drifts away from the true task execution state, leading to aimless wandering and failure to execute essential maneuvers in the instruction. 
We attribute this failure to two distinct cognitive deficits: \textit{Progress Drift}, where the agent fails to distinguish completed sub-goals from remaining ones, and \textit{Memory Drift}, where the agent's history representations degrade, making it lose track of visited landmarks. 
In this paper, we propose a Dual-Anchoring Framework that explicitly anchors the instruction progress and history representations. First, to address progress drift, we introduce Instruction Progress Anchoring, which supervises the agent to generate structured text tokens that delineate completed versus remaining sub-goals.
Second, to mitigate memory drift, we propose Memory Landmark Anchoring, which utilizes a Landmark-Centric World Model to retrospectively predict object-centric embeddings extracted by the Segment Anything Model, compelling the agent to explicitly verify past observations and preserve distinct representations of visited landmarks.
Facilitating this framework, we curate two extensive datasets: 3.6 million samples with explicit progress descriptions, and 937k grounded landmark data for retrospective verification. 
Extensive experiments in both simulation and real-world environments demonstrate the superiority of our method, 
achieving a 15.2\% improvement in Success Rate and a remarkable 24.7\% gain on long-horizon trajectories. 
To facilitate further research, we will release our code, data generation pipelines, and the collected datasets.
\end{abstract}

\section{Introduction}
\label{sec:intro}

Vision-Language Navigation~(VLN) has emerged as a central challenge in embodied artificial intelligence~\cite{duan2022survey,li2025regnav}, requiring an agent to follow natural language instructions to navigate 
in a previously unseen environment. Due to its profound potential for real-world applications in domestic service~\cite{forlizzi2006service,wisspeintner2009robocup}, intelligent personal assistants~\cite{martinez2017personal,mivseikis2020lio}, etc, this task has garnered significant attention and witnessed rapid development in recent years.

\begin{figure}[tb]
\centering
\includegraphics[width=\textwidth]{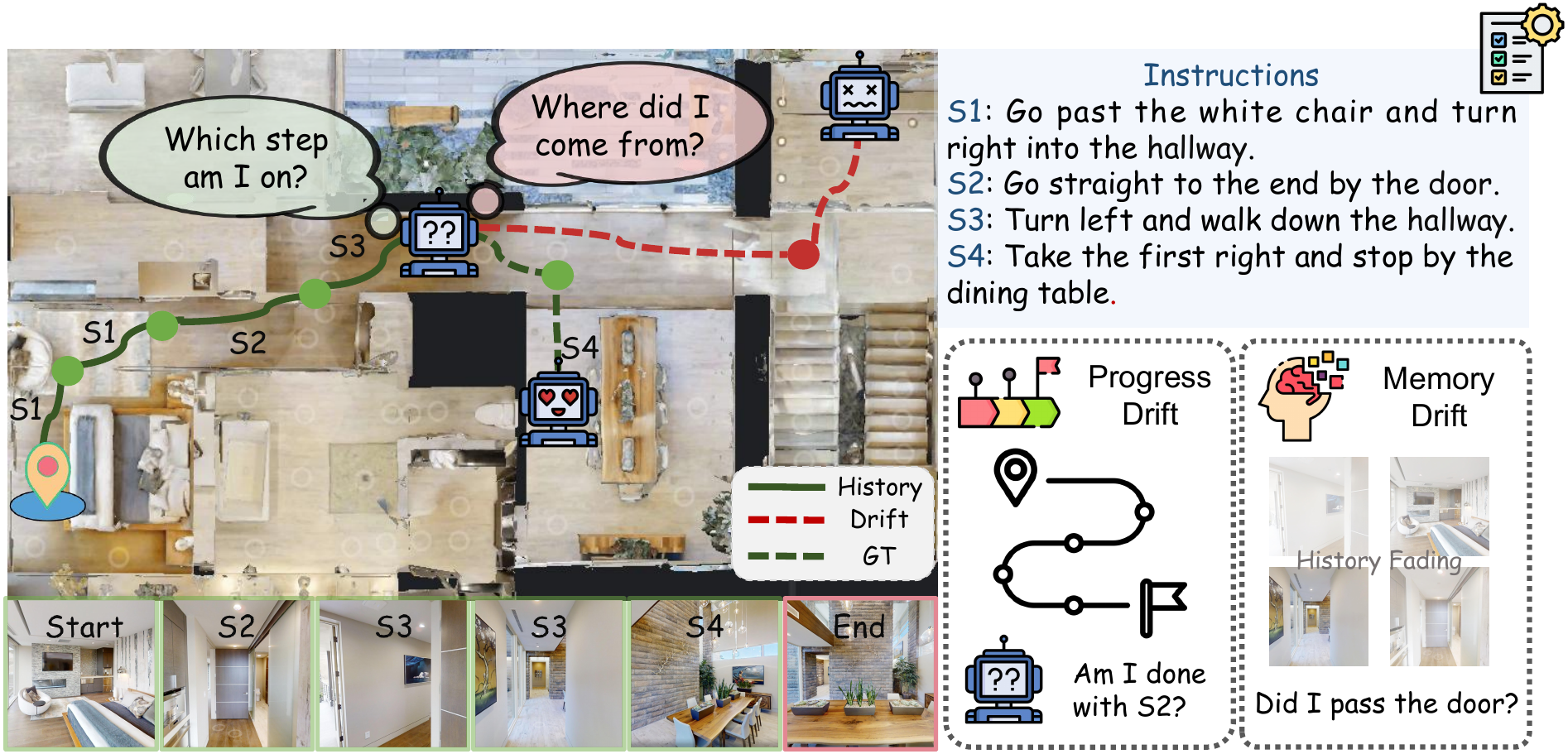}
\caption{\textbf{The Challenge of State Drift}. As the trajectory extends, the agent's predicted path deviates from the ground truth due to internal state decoupling. This manifests as the Progress Drift (confusion about which instruction step is active) and Memory Drift (failure to remember visited landmarks).}
\label{fig:intro}
\end{figure}

Most VLN works follow a perception-to-action pipeline: the agent conditions on egocentric observations and the instruction to predict an action, iterating until reaching the goal.
While effective for short instructions, this stepwise policy is brittle under long, compositional instructions. As navigation proceeds, agents may exhibit severe \textbf{State Drift}: their task state becomes progressively unreliable, and the agent gradually loses a consistent sense of (i) where it is in the instruction and 
(ii) where it has been in the environment. 
In other words, the internal state increasingly drifts away from the true task state, leading to erratic behaviors. As illustrated in Fig.~\ref{fig:intro}, we attribute State Drift to two coupled failure modes: (1) \textbf{Progress Drift}, where instruction stage is mis-tracked and the boundary between completed and remaining sub-goals becomes blurred; and (2) \textbf{Memory Drift}, where history representations degrade over long trajectories, causing the agent to lose track of visited landmarks.

Recent Video-LLMs~\cite{lin2024vila, guo2024llava} have advanced VLN~\cite{liu2025navforeseeunifiedvisionlanguageworld,internnav2025,wei2025streamvln,zhang2024navid,lyu2026himemvln,li2026spaact,li2026think} by leveraging powerful pre-trained representations. However, they remain prone to state drift in long-horizon scenarios. Fundamentally, these models rely on next-action prediction, a result-oriented objective that enforces correct local moves but neglects the agent's internal cognitive process. 
This supervision allows the agent to ``guess" the right action based on shallow local cues. However, without explicit constraints on the internal states, they will inevitably decouple from physical reality as navigation proceeds.

To address these challenges, we argue that robust navigation needs explicit state anchoring for instruction progress and visual history.  
For the former, we use structured language as a compact representation of instruction execution status. Instead of relying on implicit attention to infer progress, the agent explicitly maintains a ``mental checklist'', structurally delineating completed sub-goals from remaining ones. For the latter, we introduce a retrospective constraint: robust state grounding requires not only forward-looking control but also explicit verification of past observations. 
We therefore compel the model to preserve distinct, landmark-centric history representations to firmly anchor its current position.

In this paper, we propose the Dual-Anchoring Framework, which regularizes the Video-LLM's internal state through two complementary branches. First, to explicitly track instruction progress, we introduce \textbf{Instruction Progress Anchoring}. By designing a Progress Annotation Generation pipeline, we curate 3.6 million navigation samples where each action is prefixed with a structured progress description, explicitly delineating completed versus remaining sub-goals. Then, we utilize Instruction-Aware Co-training to force the model to articulate a concrete linguistic state before executing any control. Second, to enforce historical verification, we propose \textbf{Memory Landmark Anchoring}. To realize this, we devise a Landmark Frame Mining method to collect 937k grounded landmark data. Then, we design a Landmark-Centric World Model that operates retrospectively: 
it reconstructs the high-level semantic features of the most recently passed landmark from output tokens of history frames. 
This module acts as a "rear-view mirror," compelling the model to preserve distinct, grounded representations of its history.

Our main contributions are summarized as follows:
\begin{itemize}
\item We identify State Drift as a critical bottleneck in VLN and propose the Dual-Anchoring Framework, a unified solution that explicitly regularizes the agent's internal state, greatly eliminating erratic behaviors.
\item We introduce Instruction Progress Anchoring, backed by a scalable pipeline that synthesizes 3.6 million structured samples to enforce explicit instruction state tracking. 
Complementarily, we design a Memory Landmark Anchoring that collects 937k grounded landmark data and 
leverages SAM-based retrospective prediction to anchor the agent's visual memory.
\item We conduct extensive experiments on VLN-CE benchmarks, where our method achieves state-of-the-art~(SOTA) performance. Furthermore, real-world deployment demonstrates the framework's superior robustness and generalization in complex, unseen environments.
\end{itemize}

\section{Related Works}

\subsection{Vision Language Navigation.}
Vision-Language Navigation in Continuous Environments (VLN-CE)~\cite{krantz2020beyond} requires agents to interpret instructions and execute low-level actions in physics-realistic 3D spaces. Unlike graph-based settings~\cite{zhou2024navgpt,chen2021history,chen2022think,hong2020recurrent}, VLN-CE presents significant challenges in aligning continuous perception with long-term planning. The emergence of Video Large Language Models~(VideoLLM)~\cite{lin2024vila, guo2024llava,bai2025qwen3,bai2025qwen2} has bifurcated recent solutions into Hierarchical Planners and End-to-End Navigators. Hierarchical methods~\cite{internnav2025} decouple reasoning and control, treating the VLM as a "slow system" for mid-level plans. Conversely, End-to-End models~\cite{wei2025streamvln,zhang2024navid} fine-tune Video-LLMs to map streaming observations directly to actions, leveraging the base model's strong generalization. However, both approaches remain susceptible to state drift during extended trajectories. Without explicit state management, the agent's internal representation tends to decouple from the physical reality. In this work, we mitigate this drift by enforcing explicit regularization on the decision process: we supervise the agent to articulate structured progress descriptions for semantic alignment, and leverage an object-centric objective to predict historical features for visual grounding.
\subsection{Multi-task Training for VLN.}
Standard VLN training relies on human-annotated navigation datasets like R2R~\cite{anderson2018vision} and RxR~\cite{ku2020room}. However, the scarcity of such high-quality trajectory data limits the agent's ability to generalize across diverse unseen environments. To address this, the community has embraced Multi-task Training, broadly categorized into two paradigms. The first focuses on Broad Visual Generalization, where agents are co-trained on massive synthetic environments~\cite{internnav2025} or real-world human videos~\cite{cheng2024navila} to master diverse visual distributions. The second emphasizes Reasoning Alignment, employing auxiliary tasks like Hierarchical Planning~\cite{liu2025navforeseeunifiedvisionlanguageworld} or Chain-of-Thought (CoT) generation~\cite{huang2025mobilevla} to rationalize the perception-action loop. 
However, these auxiliary tasks remain predominantly action-centric or instantaneous, focusing primarily on interpreting the current scene for immediate decision-making. Consequently, they often neglect the explicit alignment between the accumulated history and the instruction, leaving agents vulnerable to Progress Drift. Our Instruction Progress Anchoring shifts focus from local perception to global status tracking, strictly enforcing a structural distinction between ``Done'' and ``Remaining'' parts to lock the decision process.
Similar auxiliary supervision strategies have been explored more broadly in multimodal learning, aiming to stabilize optimization and improve robustness beyond direct end-task prediction~\cite{li2023pseudo,li2024camera,liu2024semantic,liu2025mind,wu2025event,wu2025cem,lin2026morn}.
\subsection{World Model for VLN.}
World models serve as internal representations of the environment, enabling agents to perform mental simulation by predicting future states. Early approaches, such as NWM~\cite{bar2025navigation} and AstraNav-World~\cite{hu2025astranav}, operate as Generative Simulators, synthesizing high-fidelity future video streams to learn general physical priors. However, pixel-level generation is computationally prohibited for real-time onboard inference. To address this efficiency bottleneck, recent works like LS-NWM~\cite{zhang2025latent}, NavMorph~\cite{yao2025navmorph}, and NavForesee~\cite{liu2025navforeseeunifiedvisionlanguageworld} shift towards Feature-level Prediction, forecasting compact latent dynamics or semantic features instead of raw images.
However, these methods remain fundamentally future-centric. They prioritize foresight—anticipating what will happen next—while neglecting the explicit maintenance of what has already happened. In long trajectories, this lack of historical grounding leaves agents vulnerable to Memory Drift, where the history representation fades. Diverging from this foresight-dominant paradigm, we propose a Landmark-Centric World Model that prioritizes hindsight. Instead of hallucinating uncertain futures, we leverage the Segment Anything Model to reconstruct the object-centric features of previously visited landmarks. This retrospective objective compels the agent to anchor its internal state to the concrete visual history, effectively mitigating perceptual aliasing.

\section{Method}

 \begin{figure}
\includegraphics[width=\textwidth]{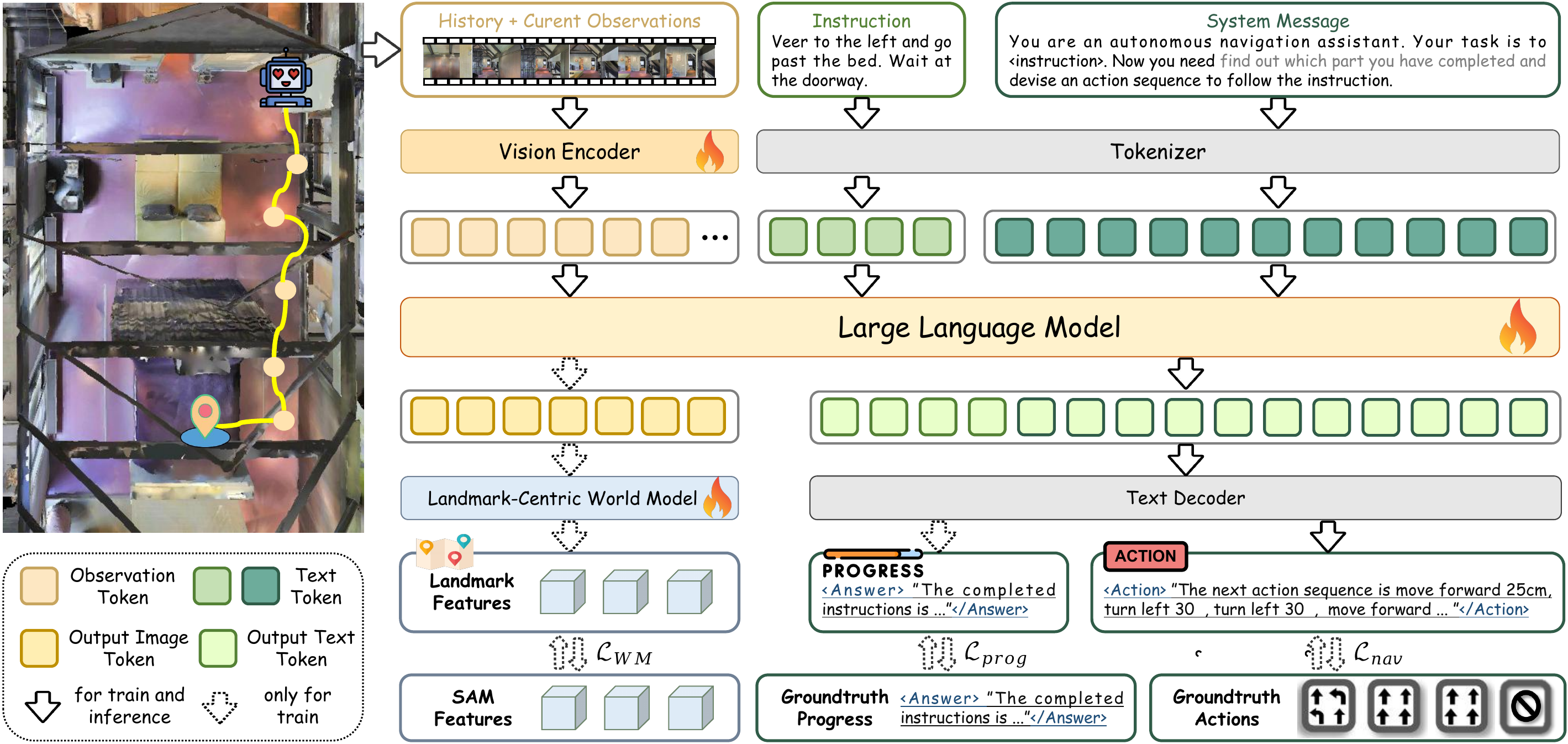}
\caption{\textbf{Overview of the Dual-Anchoring Framework.} The model processes natural language instructions and streaming egocentric observations through a unified Video-LLM backbone. Beyond standard action prediction, the framework incorporates two auxiliary tasks: generating structured progress descriptions to ensure semantic alignment, and reconstructing dense SAM features to anchor the visual memory.}
\label{fig:framwork}
\vspace{-2mm}
\end{figure}

\subsection{Problem Formulation and Overall Architecture}
\label{sec:preliminaries}

\paragraph{\textbf{Problem Formulation.}}
We formulate the Vision-Language Navigation (VLN) task as a sequential decision-making process in a continuous 3D environment. The agent is initialized at a starting position and provided with a natural language instruction $\mathcal{I} = \{w_1, w_2, \dots, w_L\}$. At each time step $t$, the agent perceives the environment via a monocular RGB observation $o_t$. Based on the instruction $\mathcal{I}$ and the history of observations, the agent predicts a low-level navigational action $a_t \in \mathcal{A}$ (e.g., \texttt{move\_forward}, \texttt{turn\_left}, \texttt{turn\_right},\texttt{stop}). Success is measured by whether the agent stops within a threshold distance of the target.

We adopt the architecture of StreamVLN~\cite{wei2025streamvln} as our backbone. This framework extends standard Video-LLMs to support streaming input by maintaining a history context $\mathcal{H}_t$ that aggregates history information.
The agent predicts the next action $a_t$ autoregressively by conditioning on the instruction and this history context:
\begin{equation}
P(a_t | \mathcal{I}, \mathcal{H}_t) = \text{VideoLLM}_{\theta}(\mathcal{I}, \mathcal{H}_t),
\label{eq:base_model}
\end{equation}
where $\theta$ is the trainable parameter. Despite the efficiency of this streaming architecture, relying solely on implicit hidden states of $\mathcal{H}_t$ to track long-horizon progress remains challenging, motivating the need for explicit state anchoring.

\paragraph{\textbf{Overall Architecture.}} 
The framework is illustrated in Fig.~\ref{fig:framwork}. To explicitly combat the State Drift dilemma in VLN, we propose the \textbf{Dual-Anchoring Framework}. While maintaining the efficient streaming backbone of StreamVLN for low-latency action generation, we augment the training architecture with Instruction Progress Anchoring and Memory Landmark Anchoring to prevent the agent's mental state from decoupling from the physical reality. The former is designed as a Co-Training Task while the latter as auxiliary head, 
thus incurring no additional computation for deployment.

\begin{figure}[tb]
\centering
\includegraphics[width=\textwidth]{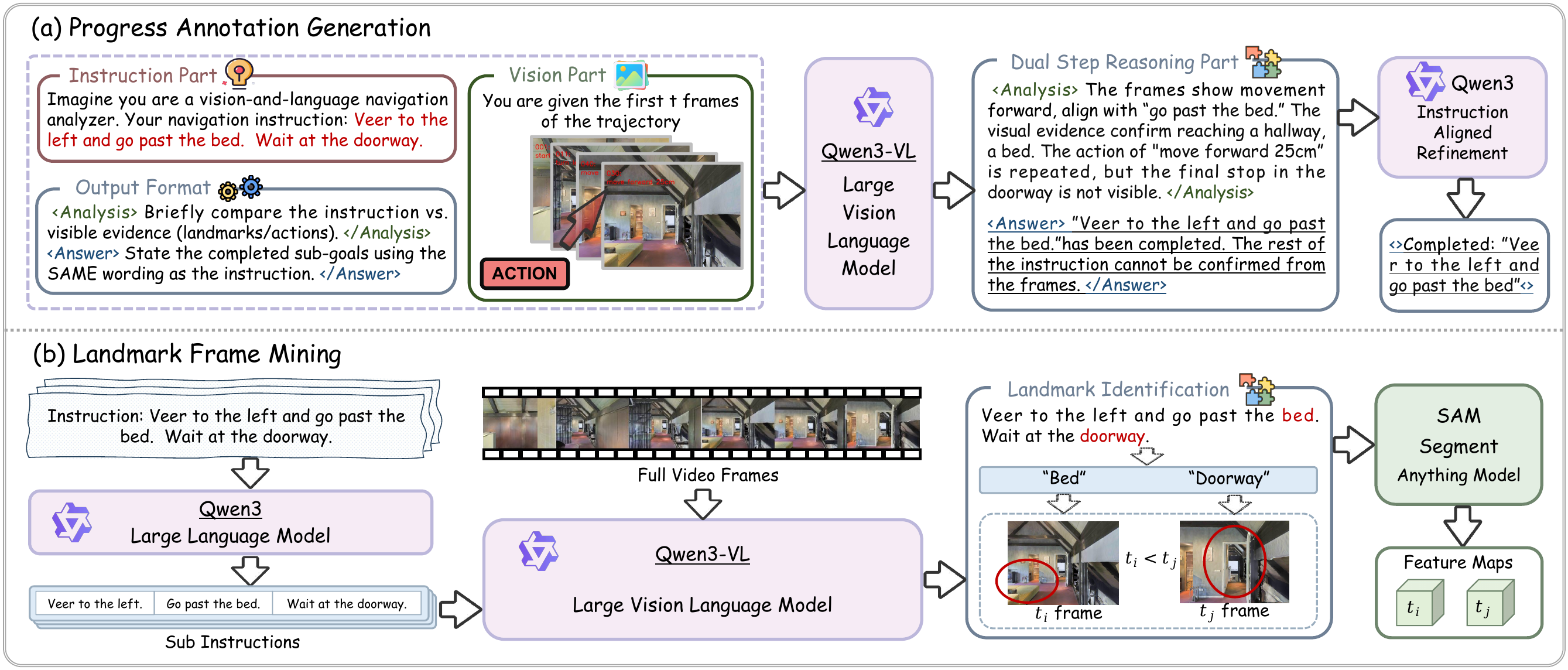}
\caption{\textbf{Overview of the data generation pipeline.} (a) A multimodal LLM performs dual-step reasoning to generate 3.6 million progress descriptions that are strictly aligned with instruction wording. (b) A hierarchical approach decomposes instructions into atomic sub-instructions and identifies temporal landmarks. The Segment Anything Model then extracts 937k object-centric features to serve as retrospective ground truth.}
\label{fig:dage_gen}
\vspace{-2mm}
\end{figure}

\subsection{Instruction Progress Anchoring}
\label{sec:ipa}

To mitigate Progress Drift, the agent must explicitly maintain a mental checklist of its execution status. However, standard VLN datasets only provide coarse-grained trajectory-level instructions, lacking step-by-step state annotations. To bridge this gap, we introduce an automated pipeline to synthesize fine-grained progress description data at scale.

\paragraph{\textbf{Progress Annotation Generation.}}
We leverage a powerful Multimodal LLM~( Qwen3-VL) as an offline annotator to generate textual progress summaries $y_t^{prog}$ given a ground-truth trajectory $\tau = \{(o_t, a_t)\}_{t=0}^{T}$ and its instruction $\mathcal{I}$. As illustrated in Fig.~\ref{fig:dage_gen}, the pipeline operates in four sequential steps.
\textit{1)~Visual Kinematics Prompting:} To bridge the gap between static frames and dynamic navigation, we explicitly overlay the frame index and executed action text~(e.g., ``Turn Left 15$^{\circ}$'') onto the top-left corner of each image frame. This explicitly informs the annotator of the agent's kinematics between timestamps.
\textit{2)~Interval Sampling:} We perform interval sampling along the trajectory with a stride of $n$. For each sampled step $t$, the annotator receives the instruction $\mathcal{I}$ and the processed visual history sequence $o_{0:t}$.
\textit{3)~Dual-Step Reasoning:} To ensure high-quality annotation, we prompt the model to perform a chain-of-thought process: first analyzing the visual evidence against the instruction
, and then generating a concise summary of \textit{Completed Sub-goals} using the original wording.
\textit{4)~Instruction-Aligned Refinement:} Finally, to eliminate intermediate reasoning and possible hallucinations, we prompt the LLM to distill the analysis into a single, concise sentence. This step strictly constrains the output to be a verbatim prefix of the instruction.
In this way, we collect a large-scale dataset comprising 3.6 million progress description samples.

\paragraph{\textbf{Instruction-Aware Co-training.}} During training, we treat this as a co-training task, where the ground-truth action is prefixed with the synthesized progress descriptions. In this case, we add ``find out which part you have completed'' in the Prompt. The generation objective at step $t$ is modified to maximize $P(y_t^{prog}, a_t | \mathcal{I}, \mathcal{H}_t)$, forcing the agent to articulate its status before action.

\subsection{Memory Landmark Anchoring}
\label{sec:world_model}

While Instruction-Aware Co-training explicitly track instruction progress, the agent still suffers from Memory Drift—forgetting distinct landmarks and falling into perceptual aliasing. Therefore, we propose Memory Landmark Anchoring that enforces \textit{Retrospective Grounding}. 

\paragraph{\textbf{Landmark Frame Mining.}}
Constructing supervision for this objective requires identifying \textit{when} specific landmarks mentioned in the instruction appear in the video. To realize this, we design a two-stage extraction process. The pipeline is illustrated in Fig~\ref{fig:dage_gen} (b). \textbf{First}, in the Decomposition stage, we prompt the LLM~(Qwen3) to decompose the complex instruction $\mathcal{I}$ into a sequence of atomic sub-goals $\mathcal{S} = \{s_1, s_2, \dots, s_K\}$, where each $s_k$ contains a specific action or landmark. \textbf{Second}, during Temporal Grounding, we feed the full video and $\mathcal{S}$ to the MLLM~(Qwen3-VL) to identify the \textit{first} appearance frame $t_{\text{lm}}^{(k)}$ for the landmark in each $s_k$. To eliminate hallucination, we enforce a strict ordering constraint: the appearance times must be strictly increasing ($t_{\text{lm}}^{(i)} < t_{\text{lm}}^{(j)}$ for $i < j$), and any annotation violating this temporal logic is filtered out. For any given time step $t$ during navigation, this process allows us to retrieve the most recently passed landmark frame index $t^*$ (where $t^* \le t$). To construct the dense supervisory signal, we extract the high-resolution spatial feature map $F_{SAM}(o_{t^*})$ from this identified frame using the Segment Anything Model (SAM)~\cite{kirillov2023segment}. This object-centric feature serves as the retrospective ground truth for our world model.

\paragraph{\textbf{Landmark-Centric World Model.}}
To effectively mitigate Memory Drift, the agent must maintain a continuous awareness of where it has been. Our World Model achieves this by compelling the agent to reconstruct the dense spatial features of the most recently passed landmark.

To realize that, we utilize a \textbf{Learnable Spatial Query Decoder}. Let $X_t \in \mathbb{R}^{N \times d_{llm}}$ denote the output sequence of the VideoLLM at step $t$, which encapsulates both historical and current visual semantics. To reduce computational overhead, we first project $X_t$ into a compact latent space via a linear layer with LayerNorm: 
\begin{equation}
\hat{X}_t = \text{LayerNorm}(X_t W_{in}) \in \mathbb{R}^{N \times d_{attn}}.
\end{equation}

Next, a set of learnable spatial queries $Q_{spa} \in \mathbb{R}^{(H \times W) \times d_{attn}}$ is initialized, where each query acts as a pixel-wise anchor corresponding to the target spatial resolution $H \times W$. These queries retrieve highly relevant local spatial cues from $\hat{X}_t$ via cross-attention:
\begin{equation}
Z = \text{Softmax}\left(\frac{Q_{spa} \hat{X}_t^T}{\sqrt{d_{attn}}}\right) \hat{X}_t \in \mathbb{R}^{(H \times W) \times d_{attn}},
\end{equation}
where $Z$ adaptively aggregates the required spatial information while preserving semantic heterogeneity. Subsequently, $Z$ is linearly projected to the target channel dimension $d_{sam}$ and reshaped into a 2D spatial format $F_t \in \mathbb{R}^{d_{sam} \times H \times W}$. 

Finally, the objective is to minimize the Mean Squared Error (MSE) between the predicted feature map $\hat{F}_t$ and the frozen SAM feature $F_{SAM}(o_{t^*})$:
\begin{equation}
\mathcal{L}_{WM} = || F_t - F_{SAM}(o_{t^*}) ||_2^2.
\end{equation}
This objective acts as a ``rear-view mirror'', compelling the agent's internal state to retain dense and distinct object information from its past trajectory, effectively preventing the memory from decaying. 

\subsection{Data Composition and Training Strategy}
\label{sec:training_strategy}

To ensure both the robustness and generalizability of our model, we compile a comprehensive training set combining standard navigation benchmarks, our synthesized anchoring data, and general vision-language data.

\paragraph{\textbf{Data Composition.}}
As summarized in Table~\ref{tab:data_composition}, our training mixture consists of four main categories. The Base Navigation data includes 180K trajectories from R2R, RxR, and EnvDrop, alongside 155K ones from a ScaleVLN subset. To explicitly enforce state anchoring, we augment these trajectories with our self-collected State Anchoring data, comprising 3.6M progress descriptions and 937K grounded landmark SAM features. Finally, the dataset is supplemented with 240K DAgger rollouts for robust policy learning, as well as 400K VideoQA pairs and 230K image-text pairs to preserve general vision-language capabilities.

\begin{table}[h]
\centering
\renewcommand{\arraystretch}{1}
\caption{\textbf{Overview of Training Data Composition and Utilization.} }
\label{tab:data_composition}
\resizebox{1\columnwidth}{!}{%
\begin{tabular}{l l c c c}
\toprule
\textbf{Dual Category} & \textbf{Source / Dataset} & \textbf{Scale} & \textbf{Objective} & \textbf{Stage} \\
\midrule
\multirow{2}{*}{Base Navigation} & R2R, RxR, EnvDrop & 180K & $\mathcal{L}_{nav}$ & 1, 2 \\
& ScaleVLN (HM3D Subset) & 155K & $\mathcal{L}_{nav}$ & 1, 2 \\
\midrule
\rowcolor{gray!10} 
 & \textbf{Progress Description Data} & \textbf{3.6M} & $\mathcal{L}_{prog}$ & \textbf{1, 2} \\
\rowcolor{gray!10}
\multirow{-2}{*}{\textbf{State Anchoring}} & \textbf{Grounded Landmark Data} & \textbf{937K} & $\mathcal{L}_{WM}$ & \textbf{1, 2} \\
\midrule
Expert Correction & DAgger Rollouts & 240K & $\mathcal{L}_{nav}$ & 2 \\
\midrule
\multirow{2}{*}{Generalist VL} & VideoQA (LLaVA-Video, ScanQA) & 400K & Next-Token & 2 \\
& Interleaved Image-Text (MMC4) & 230K & Next-Token & 2 \\
\bottomrule
\end{tabular}
}
\end{table}

\paragraph{\textbf{Two-Stage Training Pipeline.}}
We adopt a two-stage training paradigm to enhance the agent's capabilities progressively:

\paragraph{Stage 1: Navigation Data Pre-training.} The model is first trained on the aggregated navigation datasets augmented with our Dual-Anchoring objectives. The overall loss function is a weighted sum of the standard navigation action loss $\mathcal{L}_{nav}$, the progress generation loss $\mathcal{L}_{prog}$, and the world model MSE loss $\mathcal{L}_{WM}$:
\begin{equation}
\mathcal{L}_{Stage1} = \mathcal{L}_{nav} + \lambda_{prog}\mathcal{L}_{prog} + \lambda_{WM}\mathcal{L}_{WM}.
\end{equation}

\paragraph{Stage 2: DAgger and Generalist Fine-tuning.} 
To mitigate exposure bias and enhance robustness, we employ Data Aggregation (DAgger). We sample trajectories using the Stage 1 policy, collecting corrective expert actions to form a DAgger dataset of approximately 240K samples. In this stage, we co-train the model on a mixture of the navigation data from Stage 1, the newly collected DAgger data, and \textit{General Vision-Language Data} to prevent catastrophic forgetting of pre-trained knowledge. The general data includes VideoQA samples (LLaVA-Video-178K, ScanQA) and interleaved image-text data (MMC4). The Dual-Anchoring objectives ($\mathcal{L}_{prog}$ and $\mathcal{L}_{WM}$) remain strictly active for all navigation-related batches during this final fine-tuning stage.

\section{Experiments}
\label{sec:experiments}

\subsection{Experimental Setup}
\paragraph{\textbf{Datasets and Evaluation Metrics.}} We evaluate our proposed method on two continuous VLN benchmarks: \textbf{R2R-CE}~\cite{krantz2020beyond} and \textbf{RxR-CE}~\cite{ku2020room}. 
R2R-CE is based on the Matterport3D simulator and contains 90 scenes with 5.6k trajectories. 
RxR-CE is a larger dataset with more complex paths and fine-grained instructions, which is particularly suitable for evaluating the agent's ability to handle long-horizon navigation and instruction following. We report standard metrics for VLN-CE: Success Rate (\textbf{SR}), Success weighted by Path Length (\textbf{SPL}), Oracle Success Rate (\textbf{OSR}), and Navigation Error (\textbf{NE}).

\paragraph{\textbf{Implementation Details.}} Our model is built upon StreamVLN~\cite{wei2025streamvln} with LLaVA-Video~\cite{zhang2024video} as the backbone. 
We use the AdamW optimizer with a learning rate of 1e-5. 
All experiments are conducted on 32 NVIDIA H200 GPUs. Detailed dataset statistics and training hyperparameters are provided in the Appendix.

\begin{table*}[h]
\vspace{-2mm}
\centering
\caption{\textbf{Comparison with state-of-the-art methods on R2R-CE and RxR-CE validation unseen splits.} * indicates methods using the waypoint predictor from~\cite{hong2020recurrent}. The \textbf{best} and \uline{second-best} results are highlighted.}
\label{tab:main_results}
\renewcommand{\arraystretch}{1}
\resizebox{\textwidth}{!}{
\setlength{\tabcolsep}{4pt} 
\begin{tabular}{l cccc cccc c ccc}
\toprule
\multirow{2}{*}{\textbf{Method}} & \multicolumn{4}{c}{\textbf{Sensors}} & \multicolumn{4}{c}{\textbf{R2R-CE}} & & \multicolumn{3}{c}{\textbf{RxR-CE}} \\
\cmidrule(lr){2-5} \cmidrule(lr){6-9} \cmidrule(lr){11-13} 
 & \textbf{Pano.} & \textbf{Odo.} & \textbf{Depth} & \textbf{S-RGB} & \textbf{NE}$\downarrow$ & \textbf{OSR}$\uparrow$ & \textbf{SR}$\uparrow$ & \textbf{SPL}$\uparrow$ & & \textbf{NE}$\downarrow$ & \textbf{SR}$\uparrow$ & \textbf{SPL}$\uparrow$ \\
\midrule
HPN+DN*~\cite{krantz2021waypoint} & \checkmark & \checkmark & \checkmark & & 6.31 & 40.0 & 36.0 & 34.0 & & -- & -- & -- \\
CMA*~\cite{hong2022bridging} & \checkmark & \checkmark & \checkmark & & 6.20 & 52.0 & 41.0 & 36.0 & & 8.76 & 26.5 & 22.1 \\
Sim2Sim*~\cite{krantz2022sim} & \checkmark & \checkmark & \checkmark & & 6.07 & 52.0 & 43.0 & 36.0 & & 8.76 & 26.5 & 22.1 \\
GridMM*~\cite{wang2023gridmm} & \checkmark & \checkmark & \checkmark & & 5.11 & 61.0 & 49.0 & 41.0 & & -- & -- & -- \\
DreamWalker*~\cite{wang2023dreamwalker} & \checkmark & \checkmark & \checkmark & & 5.53 & 59.0 & 49.0 & 44.0 & & -- & -- & -- \\
Reborn*~\cite{an20221st} & \checkmark & \checkmark & \checkmark & & 5.40 & 57.0 & 50.0 & 46.0 & & 5.98 & 48.6 & 42.0 \\
ETPNav*~\cite{an2024etpnav} & \checkmark & \checkmark & \checkmark & & 4.71 & 65.0 & 57.0 & 49.0 & & 5.64 & 54.7 & 44.8 \\
HNR*~\cite{wang2024lookahead} & \checkmark & \checkmark & \checkmark & & 4.42 & 67.0 & 61.0 & 51.0 & & 5.50 & 56.3 & 46.7 \\
\midrule
AG-CMTP~\cite{chen2021topological} & \checkmark & \checkmark & \checkmark & & 7.90 & 39.0 & 23.0 & 19.0 & & -- & -- & -- \\
R2R-CMTP~\cite{chen2021topological} & \checkmark & \checkmark & \checkmark & & 7.90 & 38.0 & 26.0 & 22.0 & & -- & -- & -- \\
Instruc-Nav~\cite{long2024instructnav} & \checkmark & \checkmark & \checkmark & & 6.89 & -- & 31.0 & 24.0 & & -- & -- & -- \\
LAW~\cite{raychaudhuri2021language} &  & \checkmark & \checkmark &\checkmark & 6.83 & 44.0 & 35.0 & 31.0 & & 10.90 & 8.0 & 8.0 \\
CM2~\cite{georgakis2022cross} &  & \checkmark & \checkmark &\checkmark & 7.02 & 41.0 & 34.0 & 27.0 & & -- & -- & -- \\
WS-MGMap~\cite{chen2022weakly} &  & \checkmark & \checkmark & \checkmark& 6.28 & 47.0 & 38.0 & 34.0 & & -- & -- & -- \\
AO-Planner~\cite{chen2025affordances} & \checkmark &  & \checkmark & & 5.55 & 59.0 & 47.0 & 33.0 & & -- & -- & -- \\
Seq2Seq~\cite{krantz2020beyond} & & & \checkmark & \checkmark& 7.77 & 37.0 & 25.0 & 22.0 & & 12.10 & 13.9 & 11.9 \\
CMA~\cite{krantz2020beyond} &  & \ & \checkmark & \checkmark& 7.37 & 40.0 & 32.0 & 30.0 & & -- & -- & -- \\
\midrule
NaVID~\cite{zhang2024navid} & & & & \checkmark & 5.47 & 49.0 & 37.0 & 35.0 & & -- & -- & -- \\
Uni-NaVID~\cite{zhang2024uni} & & & & \checkmark & 5.58 & 53.5 & 47.0 & 42.7 & & 6.24 & 48.7 & 40.9 \\
NaVILA~\cite{cheng2024navila} & & & & \checkmark & 5.22 & 62.5 & 54.0 & 49.0 & & 6.77 & 49.3 & 44.0 \\
NavFoM~\cite{zhang2025embodied} & & & & \checkmark & 5.01 & 64.9 & 56.2 & 51.2 & & 5.51 & 57.4 & 49.4 \\
StreamVLN~\cite{wei2025streamvln} & & & & \checkmark & 4.98 & 64.2 & 56.9 & 51.9 & & 6.22 & 52.9 & 46.0 \\
DualVLN~\cite{wei2025ground} & & & & \checkmark & \textbf{4.05} & \textbf{70.7} & \uline{64.3} & \uline{58.5} & & \uline{4.58} & \uline{61.4} & \uline{51.8} \\
\midrule
\textbf{Ours} & & & & \checkmark & \uline{4.15} & \uline{69.2} & \textbf{65.6} & \textbf{62.1} & & \textbf{4.42} & \textbf{61.7} & \textbf{53.3} \\
\bottomrule
\end{tabular}
}
\end{table*}
\vspace{-5mm}
\subsection{Experiments on Simulated Environments}
\paragraph{\textbf{Comparison with State-of-the-Arts.}}
As summarized in Table~\ref{tab:main_results}, our Dual-Anchoring Framework establishes a new state-of-the-art on both benchmarks. On R2R-CE, our method significantly outperforms the strong StreamVLN~\cite{wei2025streamvln} baseline, improving Success Rate (SR) from $56.9\%$ to $65.6\%$ and SPL from $51.9\%$ to $62.1\%$. This demonstrates that while implicit hidden states in prior Video-LLMs tend to decouple from physical reality, our explicit regularization effectively synchronizes the agent's cognitive process with the environment. For the more challenging RxR-CE dataset, which features longer trajectories and complex instructions, the benefits of our approach are even more pronounced. While long horizons typically exacerbate State Drift, our framework maintains robustness, achieving an $8.8\%$ absolute gain in SR ($52.9\% \rightarrow 61.7\%$) over StreamVLN. These results confirm that anchoring internal states to both instruction progress and visual landmarks is crucial in long-horizon navigation.

\begin{figure}[t]
  \centering
  \includegraphics[width=\linewidth]{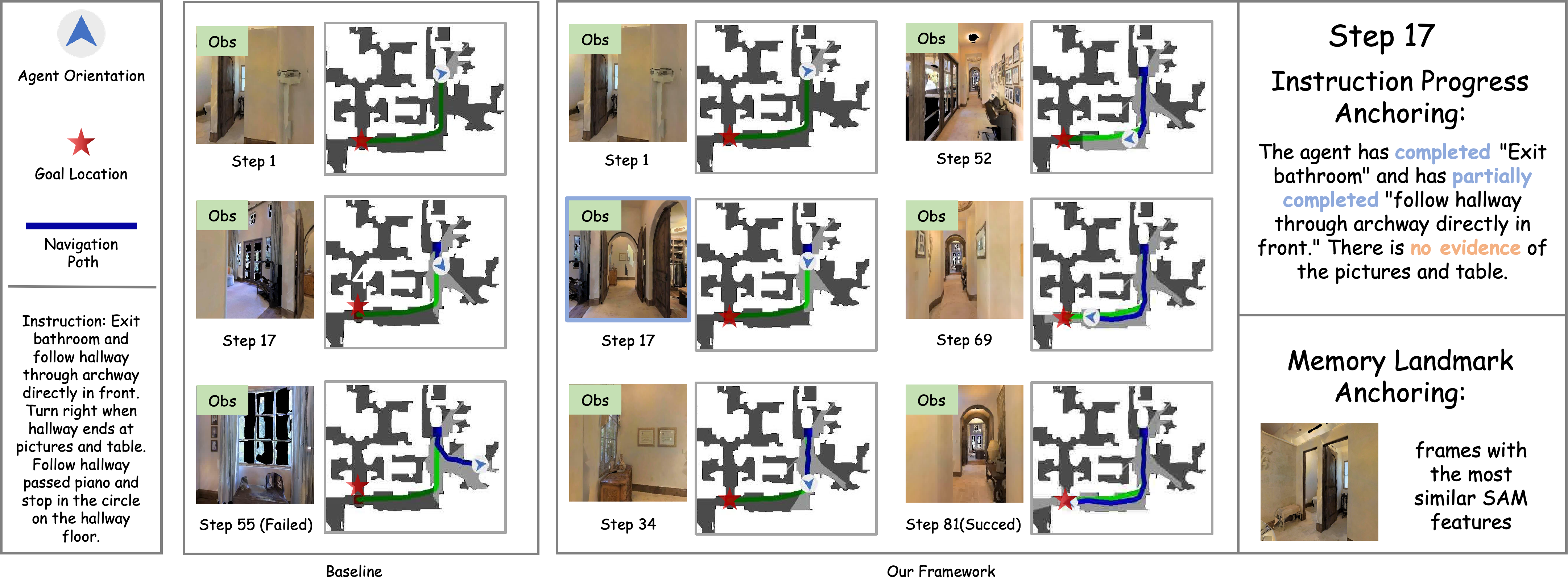}
  \caption{\textbf{Qualitative Visualization in Simulated Environment.} We compare the navigation trajectories of the Baseline and our model on an R2R-CE unseen episode. While the baseline turns prematurely due to state drift, our model maintains the correct heading. The right panel showcases the internal states of our agent: the Instruction Progress Anchoring correctly identifies the current sub-goal as only "partially completed" , and the Memory Landmark Anchoring successfully reconstructs the visual features of the previously passed "bathroom" to ground the current spatial context.  }
  \label{fig:qualitative_sim}
  \vspace{-2mm}
\end{figure}

\paragraph{\textbf{Visualization.}}
To demonstrate how our framework mitigates State Drift, we visualize a challenging R2R-CE episode in Fig.~\ref{fig:qualitative_sim}. The instruction requires the agent to "\textit{exit the bathroom, go straight to the end of the hallway, and turn left.}" 
At Step 17, when encountering a deceptive opening before the hallway's true end, the baseline suffers from Progress Drift, which assumes the navigation has reached the turning point, leading to a premature turn left and ultimate failure. 
In contrast, our Dual-Anchoring agent maintains a synchronized internal state. 
Although the auxiliary heads are typically discarded during inference to minimize latency, we employ them here as probes to decode and visualize the agent's latent consistency.
The Instruction Progress Anchoring acts as a semantic stabilizer, explicitly signaling that the "go straight" phase is still in progress, which effectively suppresses the distractor action. Simultaneously, the Memory Landmark Anchoring provides a "rear-view mirror" effect, retrospectively grounding the agent by retrieving the starting bathroom's features. This suggests that explicit state anchoring could prevent the decision-making process from decoupling from physical reality, ensuring robust long-horizon execution.

\begin{table}[h]
\vspace{-2mm}
\centering
\renewcommand{\arraystretch}{1.2}
\caption{\textbf{Ablation study of key components on R2R-CE Val Unseen.} We evaluate the Instruction Progress Anchoring~(short for IPA) and Memory Landmark Anchoring~(short for MLA) under two training data scales. To ensure fairness, the number of navigation samples remains identical across variants within the same setting.}
\label{tab:ablation}
\setlength{\tabcolsep}{4pt}
\resizebox{\columnwidth}{!}{%
\begin{tabular}{cc|cccc|cccc}
\toprule
\multicolumn{2}{c|}{\textbf{Components}} & \multicolumn{4}{c|}{\textbf{R2R+RxR+EnvDrop}} & \multicolumn{4}{c}{\textbf{All Datasets}} \\
\cmidrule(lr){1-2} \cmidrule(lr){3-6} \cmidrule(lr){7-10}
\textbf{IPA} & \textbf{MLA} & \textbf{NE} $\downarrow$ & \textbf{OS} $\uparrow$ & \textbf{SR} $\uparrow$ & \textbf{SPL} $\uparrow$ & \textbf{NE} $\downarrow$ & \textbf{OS} $\uparrow$ & \textbf{SR} $\uparrow$ & \textbf{SPL} $\uparrow$ \\
\midrule
~ & ~ & 6.49 & 46.9 & 40.8 & 37.4 & 4.98 & 64.2 & 56.9 & 51.9 \\
\checkmark & ~ & 6.27 & 51.9 & 45.4 & 41.1 & 4.72 & 66.5 & 60.8 & 55.1 \\
~ & \checkmark & 6.01 & 53.6 & 47.7 & 43.3 & 4.51 & 67.0 & 62.5 & 57.3 \\
\checkmark & \checkmark & \textbf{5.73} & \textbf{55.8} & \textbf{49.5} & \textbf{44.9} & \textbf{4.15} & \textbf{69.2} & \textbf{65.6} & \textbf{62.1} \\
\bottomrule
\end{tabular}
\vspace{-5mm}
}
\end{table}
\paragraph{\textbf{Ablation Study of Components.}}
To investigate the effect of our proposal, we conduct ablation studies on the R2R-CE validation unseen split under two training settings: 1) Standard Mixture (w/o dagger and VL data) to isolate methodological gains, and 2) All Datasets to validate robustness at scale. 
In Standard Mixture setting, we make the total number of training samples strictly identical. For example, if Instruction Progress Anchoring is added, we replace half of the base nav data in the baseline with our augmented Progress Description Data, adding no additional samples.
In All Datasets setting, we don't control the number of training samples. Therefore, if Instruction Progress Anchoring is added, we add all Progress Description Data to the original training dataset.

As shown in Table~\ref{tab:ablation}, the baseline yields the worst results due to State Drift. Incorporating the Instruction Progress Anchoring significantly improves Success Rate (e.g., from 40.8\% to 45.4\%), confirming that explicit sub-goal prediction maintains semantic alignment. Meanwhile, the Memory Landmark Anchoring notably boosts SPL and reduces Navigation Error (6.49 $\to$ 6.01), validating that retrospective landmark reconstruction effectively prevents trajectory drift. Finally, the Dual-Anchoring configuration achieves the best performance across both settings, demonstrating that our explicit state regularizations offer fundamental, complementary benefits that persist even with scaled-up data.

\paragraph{\textbf{Analysis across Trajectory Lengths.}}
\begin{figure}[h]
  \vspace{-8mm}
  \centering
\includegraphics[width=\linewidth]{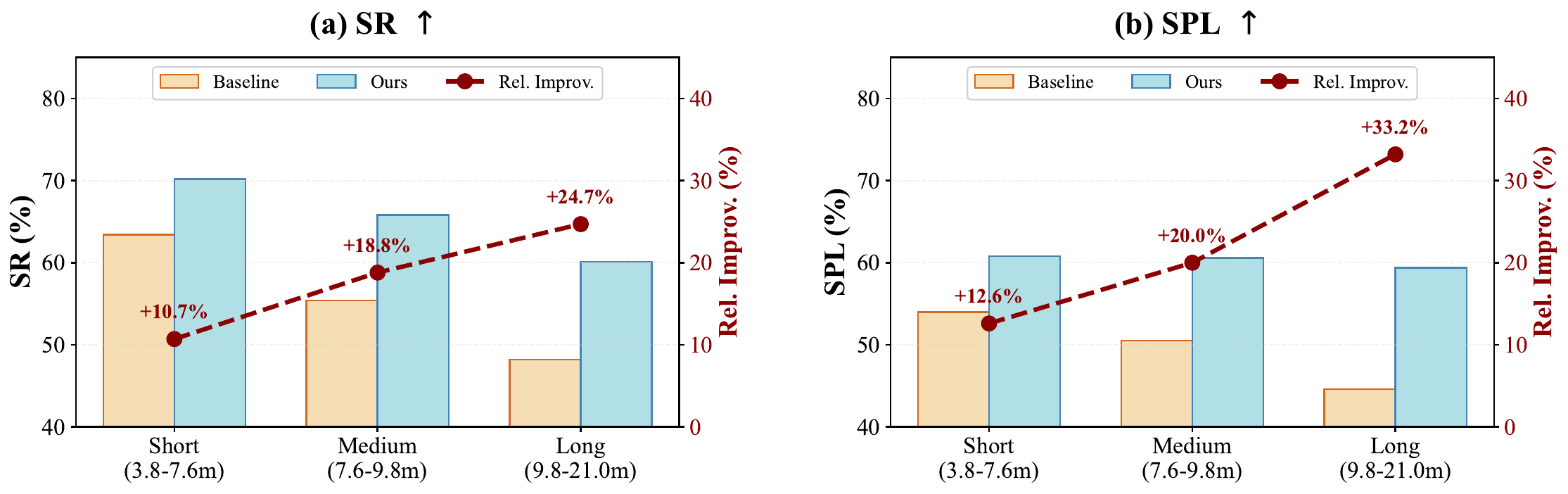}
  \caption{\textbf{Performance across different trajectory lengths.} We report (a) SR and (b) SPL across Short, Medium, and Long episodes  on R2R-CE validation unseen set. As trajectories become longer, our method provides increasingly substantial gains.}
  \label{fig:long_horizon_dist}
  \vspace{-2mm}
\end{figure}

We further analyze the performance across different trajectory lengths. The R2R-CE validation unseen set is divided into three subsets based on the trajectory geodesic distance: \textbf{Short} ($\in [3.85, 7.55)$ meters), \textbf{Medium} ($\in [7.55, 9.81)$ meters), and \textbf{Long} ($\in [9.81, 21.04]$ meters). As illustrated in Fig.~\ref{fig:long_horizon_dist}, the baseline exhibits sharp degradation as navigation distance increases due to unmitigated drift. In contrast, our Dual-Anchoring agent maintains superior stability, with the performance gap widening significantly on longer trajectories. Especially for the relative improvement in SPL, which escalates from +12.6\% in Short episodes to a remarkable \textbf{+33.2\%} in Long episodes. This trend confirms that explicit state anchoring—via both Instruction Progress and Memory Landmark modules—effectively counteracts the decoupling of internal states from physical reality, a benefit that becomes increasingly critical n long-horizon navigation scenarios.

\definecolor{bettergreen}{RGB}{34,139,34}

\begin{table}[h]
\vspace{-2mm}
\centering
\caption{\textbf{Quantitative Assessment of Selected Datasets.} We evaluate the quality of auto-generated data using reference-free metrics. The ``Baseline'' column denotes the control settings: raw image inputs (w/o Visual Prompting) for Progress Description Data, and Random Sampling for Grounded Landmark Data.}
\label{tab:data_quality}
\resizebox{\linewidth}{!}{
\begin{tabular}{l|c|l|c|cc}
\toprule
\textbf{Datasets } & \textbf{Baseline} & \textbf{Metric} & \textbf{Baseline} & \textbf{Ours} & $\bm{\Delta}$ \\
\midrule
\multirow{2}{*}{\textbf{Progress Description}} & \multirow{2}{*}{\shortstack[c]{w/o Visual\\ Prompting}} & Hallucination Rate (HR) (\%) $\downarrow$ & 8.13 & \textbf{6.04} & \textcolor{bettergreen}{\textbf{-25.7\%}} \\
 & & Logical Consistency Score (LCS) (1-5) $\uparrow$ & 1.71 & \textbf{4.26} & \textcolor{bettergreen}{\textbf{+149\%}} \\
\midrule
\textbf{Grounded Landmark} & Random Sampling & Landmark Presence Rate (LPR) (\%) $\uparrow$ & 13.9 & \textbf{75.6} & \textcolor{bettergreen}{\textbf{+444\%}} \\
\bottomrule
\end{tabular}
}
\vspace{-2mm}
\end{table}

\subsection{Quality Assessment of Self-Collected Dataset.}
Since our framework relies on self-collected datasets, ensuring their reliability is paramount. We establish a rigorous evaluation protocol to quantify the quality of Progress Description Data and Grounded Landmark Data.

\paragraph{\textbf{Progress Description Data.}}
We employ an LLM-as-a-Judge paradigm to evaluate generated progress descriptions using two key metrics:
1) \textit{Hallucination Rate (HR):} The percentage of descriptions containing entities not present in the original instruction.
2) \textit{Logical Consistency Score (LCS):} A 1-5 Likert scale measuring whether the generated progress represents a logically contiguous prefix of the instruction without skipping steps. We calculate these under two distinct settings: feeding raw image inputs VS employing our \textit{Visual Kinematics Prompting} (Sec.~\ref{sec:ipa}). As shown in Table~\ref{tab:data_quality}, the setting without visual prompting struggles to track the agent's dynamic movement, resulting in a low LCS of 1.71. In contrast, our method achieves a high LCS of \textbf{4.26} and a significantly reduced HR from {8.13\%} to {6.04\%}. This improvement confirms that overlaying explicit action history is critical for the annotator to align visual changes with instruction steps, thereby yielding high-quality, coherent, and faithful training data.

\paragraph{\textbf{Grounded Landmark Data.}}
To evaluate the temporal precision of the grounded landmark, we propose the Landmark Presence Rate (LPR), which extracts the landmark entity from the sub-instruction and prompts Qwen3-VL to verify whether the object is visually recognizable. We compare our mined frames against a Random Sampling baseline.
As reported in Table~\ref{tab:data_quality}, random frames contain the target landmark only 13.9\% of the time. Conversely, our method achieves an LPR of 75.6\%. This substantial gap confirms that our pipeline effectively localizes the semantic ``high-light'' moments of landmarks, thus providing valid retrospective signals.

\subsection{Real-World Deployment.}
\begin{figure}[tb]
\centering
\includegraphics[width=0.9\textwidth]{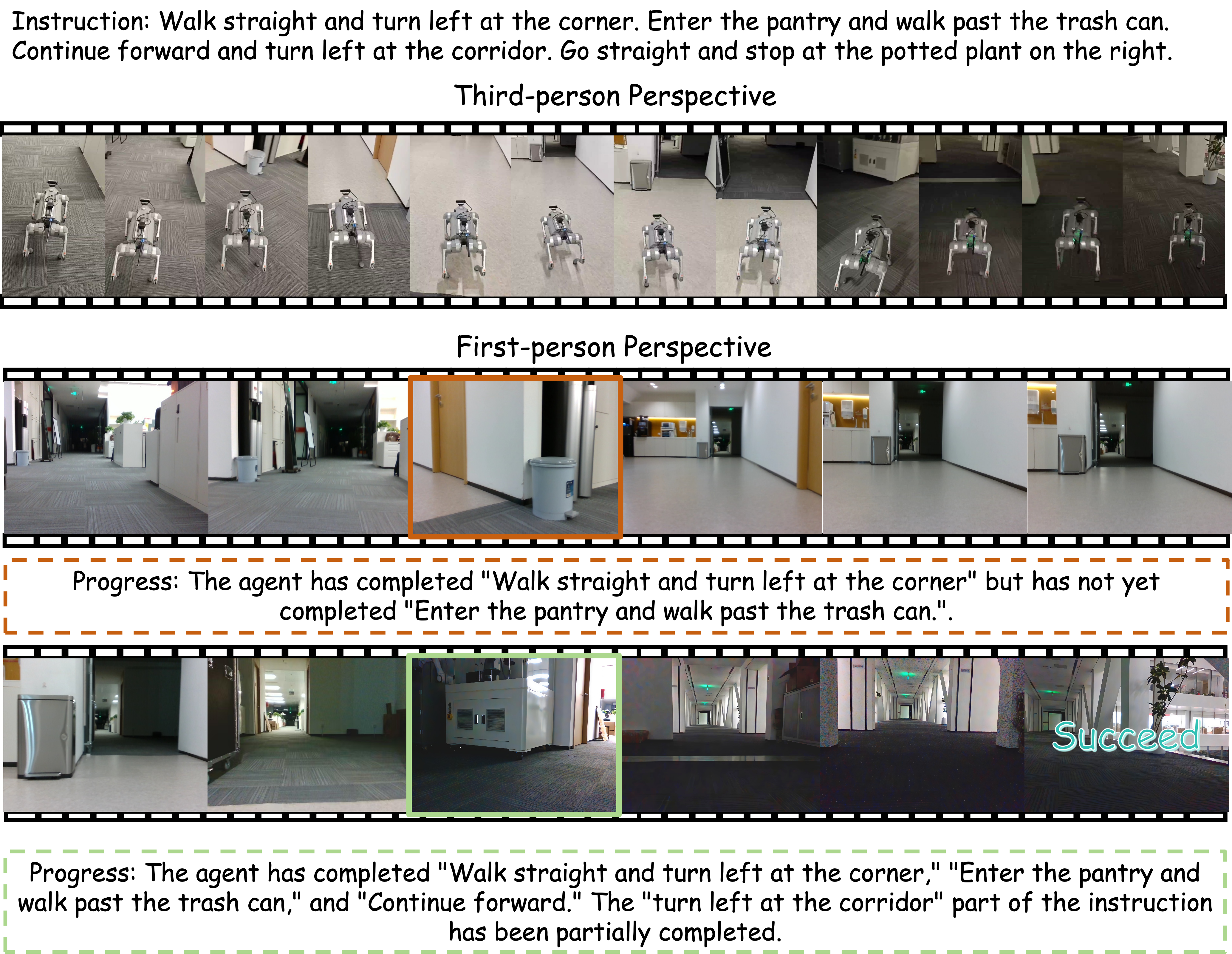}
\caption{\textbf{Qualitative Visualization of Real-World Deployment.} Top: Third-person perspective of the robot's trajectory. Bottom: First-person RGB stream. The \textbf{dashed boxes} display the real-time progress descriptions explicitly generated by our model, corresponding to the frames bordered in the same color.
}
\label{fig:realworld_qualitative}
\end{figure}

\noindent\textbf{Experimental Settings.}
Our real-world experiments are conducted on a Unitree Go2 quadruped robot equipped with a front-facing Intel RealSense D435i RGB-D camera.
The navigation system operates in a client-server architecture: RGB observations are continuously streamed from the D435i camera to a remote server powered by NVIDIA H20 GPUs.
Upon receiving a natural language instruction, our model processes the streaming observations to predict the next low-level navigational action in real-time.
The predicted action is then transmitted back to the Go2 robot, which executes it by calling the robot motion API.
Note that the model is trained entirely in the Matterport3D simulator and deployed directly to the real world without any fine-tuning.

\noindent\textbf{Qualitative Results.}
As illustrated in Fig.~\ref{fig:realworld_qualitative}, our agent successfully anchors its visual observations to linguistic sub-goals despite the significant domain gap. The model explicitly generates progress descriptions (shown in dashed boxes) that accurately reflect its current status. For instance, in the intermediate stage (\textcolor{orange}{orange box}), the agent correctly identifies that it has completed the ``turn left'' action but still needs to ``enter the pantry''. Similarly, in the near-goal stage (\textcolor{green}{green box}), it precisely recognizes the completion of the full instruction. This capability to distinguish between completed and remaining segments effectively mitigates progress drift in long-horizon tasks.

\section{Conclusion}
In this paper, we address State Drift in Vision-Language Navigation by proposing the Dual-Anchoring Framework, which explicitly anchors the agent's internal state to instruction progress and visual landmarks. To support this, we introduce Instruction Progress Anchoring and Memory Landmark Anchoring, backed by automated pipelines that generate 4.5 million synthesized samples. Extensive experiments show our method achieves state-of-the-art performance on R2R-CE and RxR-CE, with remarkable robustness in long-horizon trajectories and successful real-world deployment.
\section*{Acknowledgements}
This work was supported by the National Key R\&D Program of China (2024YFB3309303).
\clearpage  

\bibliographystyle{splncs04}
\bibliography{main}
\end{document}